\documentclass{article}

\usepackage{booktabs}
\usepackage{tabularx}
\usepackage{array}
\usepackage[table]{xcolor}
\usepackage{amssymb}
\usepackage{amsmath}
\usepackage{caption}
\usepackage{subcaption}
\usepackage{graphicx}
\usepackage{comment}

\usepackage[preprint]{corl_2026} % Uncomment for pre-prints (e.g., arxiv); This is like ``final'', but will remove the CORL footnote.

\graphicspath{{figures/}}

\title{Near-Zero Forgetting Is Not Enough: Out-of-Distribution Robustness in Continual Learning of Pretrained VLAs}

\author{
  Adrià Ruiz\\
  KU Leuven\\
  Computer Vision Center \\
  Universitat Aut\`{o}noma de Barcelona \\
  \texttt{adriarp03@gmail.com} \\
  \And
  Bartłomiej Twardowski \\
  Computer Vision Center \\
  Universitat Aut\`{o}noma de Barcelona \\
  IDEAS Research Institute \\
  \texttt{btwardowski@cvc.uab.es} \\
  \AND
  Tinne Tuytelaars \\
  KU Leuven \\
  \texttt{tinne.tuytelaars@kuleuven.be} \\
  \And
  Sergi Masip \\
  KU Leuven \\
  \texttt{sergi.masipcabeza@kuleuven.be} \\
}

\begin{document}
\maketitle

%===============================================================================

\begin{abstract}
    Recent work shows that pretrained Vision-Language-Action (VLA) models exhibit near-zero forgetting in specific continual learning settings with experience replay. However, robustness to domain shift of learned tasks remains unexplored. In this work, we study the robustness of Pi0 fine-tuned with experience replay under out-of-distribution (OOD) data using the LIBERO and LIBERO-Plus benchmarks. While the in-domain (ID) setup achieves near-zero catastrophic forgetting, OOD performance on previous tasks quickly collapses. Specifically, we show that during the continual learning process, the OOD capabilities of the model degrade, following a shared representational pattern across domain shifts, which is also visible behaviorally in the two simpler suites. Representational analysis shows that drift concentrates in the multimodal and action features. Linear probes indicate that OOD features still retain task-relevant information, but appear unreadable, consistent with representation misalignment. These findings suggest that even when ID stability remains intact, severe OOD forgetting can still occur, accompanied by representational drift during continual learning.
\end{abstract}


\section{Introduction}
\label{sec:introduction}

Vision-Language-Action models have emerged as the generalist approach for robot manipulation tasks. These models combine a pretrained Vision-Language Model (VLM) with an action head. This structure allows the transfer of generalist semantic priors to robot control tasks, enabling powerful generalization capabilities across different embodiments, tasks, and environments \citep{black2025pi_0, bjorck2025gr00t, kim2025openvla}. To avoid the costly retraining of these policies in changing real-world scenarios, continual adaptation would be desirable. This is the domain of continual learning (CL)~\citep{van2019three}, where new tasks or domains are learned sequentially. One of the main challenges is catastrophic forgetting~\citep{mccloskey1989catastrophic}, where the performance of previous tasks degrades as new ones are learned. This is closely related to the stability-plasticity trade-off~\citep{hadsell2020embracing}, which presents the tension between the model's ability to learn new knowledge (plasticity) and retain previous knowledge (stability).

While the broader CL community has already observed that large-scale pretraining can alleviate forgetting~\citep{wang2022dualprompt, zhou2025proof, biderman2024lora}, recent work on continual learning of VLA models has reported surprisingly strong resistance to forgetting in different regimes by using experience replay (ER) or naïve LoRA fine-tuning~\citep{liu2026pretrained, zhu2026, hu2026simple}. In some cases, forgetting is reduced to nearly zero or even replaced by backward transfer. If these results extend to deployment settings, they would represent a significant shift in what continual learning can achieve, potentially bringing lifelong robotic systems closer to practical deployment under changing environments.

However, while these studies focus on in-domain (ID) retention, how out-of-distribution (OOD) robustness evolves remains open. This is particularly relevant given that recent benchmarks~\citep{fei2025liberoplus, zhou2025liberopro, kim2026liberopara, wang2026liberox} have demonstrated that VLA performance often degrades when pushed beyond standard benchmark setups. The near-zero forgetting results in \citet{liu2026pretrained} and \citet{zhu2026} were obtained with experience replay, where the replay buffer accumulates demonstrations exclusively from the training distribution. This risks narrowing the policy's representations, which could severely degrade its capacity to generalize. This concern motivates the central hypothesis of this work: near-zero forgetting under ER may be specific to ID evaluation and not extend to domain shifts of already-learned tasks.

To test this hypothesis, we reproduce the \citet{liu2026pretrained} baseline with Pi0~\citep{black2025pi_0} across all four LIBERO suites, confirming their in-domain results. We then reevaluate the learned LIBERO tasks on LIBERO-Plus~\citep{fei2025liberoplus}, defining an OOD evaluation procedure that lets us assess how the model reacts to distribution shifts. Finally, we analyze the model's internal representations to better understand its OOD behavior. Our main findings are summarized as follows:

\begin{enumerate}
    \item \textbf{Despite ID near-zero forgetting, the model incurs heavy OOD forgetting.} We evaluate on OOD scenes, where we observe that the different perturbations induce a widening gap in performance between ID and OOD as the training progresses. This suggests that the model forgets its previous OOD capabilities (Section~\ref{sec:ood_forgetting}).
    \item \textbf{Failure structure is shared across domain shifts.} The structural coherence of failures across all perturbations is consistent with a shared representational bottleneck rather than perturbation-specific failures. We also find that most representational drift occurs in the multimodal and action-expert layers, whereas the vision encoder is comparatively robust (Section~\ref{sec:architectural_bottleneck}).
    \item \textbf{OOD features remain informative but misaligned.} Our analysis suggests that the representations from OOD episodes still keep features useful for planning. However, they appear misaligned with the ID representations, so an ID-trained readout cannot read them (Section~\ref{sec:misalignment}).
\end{enumerate}

\section{Related Work}
\label{sec:related_work}

\paragraph{Near-zero forgetting in pretrained VLAs.} Catastrophic forgetting is one of the main challenges in continual learning~\citep{mccloskey1989catastrophic, kirkpatrick2017overcoming}. Large-scale pretraining has been observed to mitigate forgetting beyond robotics, such as in pretrained vision backbones~\citep{wang2022learning, wang2022dualprompt}, VLMs~\citep{zhou2025proof}, or parameter-efficient fine-tuning~\citep{biderman2024lora}. Recent work pushes this further for large pretrained VLAs, reporting that they show greater resistance to catastrophic forgetting with simple strategies, and obtaining near-zero forgetting or even backward transfer~\citep{liu2026pretrained, hu2026simple, zhu2026}.

\paragraph{Domain shift under fine-tuning and continual learning.} Fine-tuning can erode OOD robustness; benchmarks such as ImageNet-RIB~\citep{hwang2025large} show that the more diversely pretrained models suffer larger losses in OOD generalization when fine-tuned. In the intersection of continual learning with domain shift approaches for evolving domains, CTDG~\citep{cai2024ctdg} models continually shifting distributions, and continual test-time adaptation~\citep{wang2022continual} updates models as the environment changes. For robotics, related work addresses domain shift in navigation~\citep{hisada2026flatmanifold} and visually robust or online adaptation of VLAs~\citep{peng2026ignore, wang2026road, kachaev2026dont}. In these works, the central question is whether the model can perform in a world with continual changes, but they do not question whether already known tasks remain robust under distribution shift of their already-learned knowledge.

\paragraph{Robustness evaluation for VLAs.} Many datasets and benchmarks exist for robotic evaluation~\citep{o2024open,walke2023bridgev2,khazatsky2024droid,gu2023maniskill,mees2022calvin} but for the specific case of continual learning, LIBERO~\citep{liu2023libero} remains the gold standard. In recent years, many works have extended the LIBERO framework to further stress-test VLAs in different aspects such as probing memorization, language conditioning or visual robustness~\citep{fei2025liberoplus, zhou2025liberopro, kim2026liberopara, wang2026liberox, fang2026vision}. Specifically, our approach uses LIBERO-Plus~\citep{fei2025liberoplus}, which maintains the original LIBERO tasks but expands into different controlled domain perturbations, showing a clear lack of generalization for different popular models.

\paragraph{Positioning.} The preceding sections show a gap at the intersection of several active research areas. The empirical reports on near-zero forgetting~\citep{liu2026pretrained, hu2026simple, zhu2026} on pretrained VLAs open the door for new boundaries in continual learning, but none of these studies account for the out-of-distribution robustness of the trained tasks. While some CL mitigation strategies for VLAs, such as ExpReS-VLA~\citep{syed2026expresvla}, report OOD evaluations, these remain adaptation scenarios rather than asking how pretrained VLAs interact with unseen data. This paper addresses this gap by studying whether near-zero forgetting extends to out-of-distribution evaluation in the experience-replay regime.
\section{Experimental Setup}
\label{sec:setup}

\subsection{Continual Learning Protocol}

We formalize a task suite $S=\{T_0,\ldots,T_{N-1}\}$ consisting of $N$ tasks that the model learns sequentially; after training on task $T_k$, we save checkpoint $\theta_k$. Each suite has an in-domain (ID) evaluation set $\mathcal{D}_S$ and a family of out-of-distribution (OOD) versions of the same tasks, $\mathcal{P}_S=\{P_S^{(1)},\ldots,P_S^{(M)}\}$, corresponding to the $M$ perturbation axes applied to that suite. An evaluation cell $C(k,i,p)$ is the evaluation of checkpoint $\theta_k$ on task $T_i$ under domain $p\in\{\mathcal{D}_S\}\cup\mathcal{P}_S$.

In all our experiments, we follow the setup in~\citep{liu2026pretrained} for Pi0~\citep{black2025pi_0}, which combines a pretrained PaliGemma VLM backbone with a flow matching action expert. We fine-tune the model using a replay buffer that accumulates 1,000 frames per task. For each subsequent task $T_{k+1}$, training batches contain a 1:1 ratio of current-task and replay samples.

\subsection{Benchmarks}

For ID evaluation, we use the four task suites from the LIBERO~\citep{liu2023libero} benchmark: LIBERO-Spatial, LIBERO-Object, LIBERO-Goal, and LIBERO-10. To measure domain shift, we use LIBERO-Plus~\citep{fei2025liberoplus}, which extends these suites with different domain perturbations applied to the original LIBERO tasks. Specifically, LIBERO-Plus introduces seven perturbation axes that include changes to the background textures, the robot initial state, the external camera viewpoint, the layout of the objects, the light conditions, the rewording of the language instructions, and the addition of sensor noise.

\subsection{Evaluation Metrics}

To evaluate continual learning performance, we rely on different metrics used in the continual learning literature. We compute the average cumulative accuracy (ACA), which measures the performance on all the tasks the model has seen up to the current checkpoint:
\begin{equation}
    ACA_k = \frac{1}{k+1} \sum_{i=0}^k SR(k, i, p).
\end{equation}
Intuitively, a reduction in the ACA translates to a decrease in the general understanding of the tasks.

To measure forgetting and, therefore, stability, we use negative backward transfer (NBT) as described by \citet{liu2023libero}. This metric tracks performance drops relative to when the task was initially learned. For $N\ge2$, with $NBT_{N-1}=0$, it is defined as follows:
\begin{equation}
    NBT = \frac{1}{N} \sum_{i=0}^{N-1} NBT_i, \quad
    NBT_i = \frac{1}{N - i - 1} \sum_{k=i+1}^{N-1} \Big( SR(i, i, p) - SR(k, i, p) \Big).
\end{equation}

A low NBT means the model is highly stable and does not forget past tasks, while a negative NBT implies backward transfer. We compute it on both ID and OOD domains $p$. We use OOD NBT to quantify OOD forgetting.

\subsection{Representation Analysis}

We analyze the evolution of the model's internal representations at three feature-extraction points: after the vision encoder, after the VLM, and after the first flow matching step at the action module. All features are taken from a single joint forward pass with the flow-matching step locked at $\tau = 0$, and then mean-pooled over non-padding tokens into one vector per frame. These vectors are then used to compare across checkpoints and domains. We use Linear CKA~\citep{kornblith2019similarity} to compute the drift in the same tasks (lower CKA implies more drift), Maximum Mean Discrepancy~\citep{gretton2012kernel} to measure the distributional gap between ID and OOD (higher MMD implies a larger gap), and linear probes~\citep{alain2016understanding} to quantify how much task-relevant information in the representations is linearly accessible (lower~$R^2$ equals less readable). Details are in Appendix~\ref{app:repr_settings}.
\section{Near-Zero In-Domain Forgetting Does Not Extend to OOD}
\label{sec:ood_forgetting}

\begin{figure}[t]
    \centering
    \includegraphics[width=\linewidth]{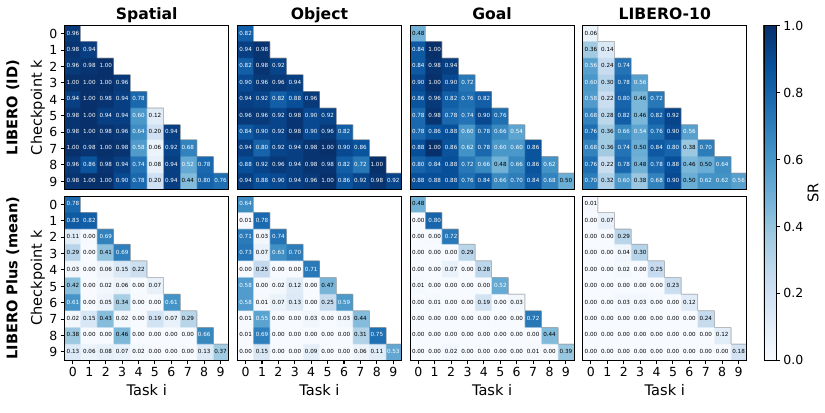}
    \caption{\textbf{ID and OOD success rate matrices for LIBERO and LIBERO-Plus (mean across perturbation categories).} Each cell
    $C(k,i)$ reports the success rate of checkpoint $\theta_k$ (row) on task
    $T_i$ (column); only valid cells with $k \geq i$ are shown. For the first row of subplots, the ID cells are shown; for the second, the mean of the cells across the OOD categories is shown. For ID, the dense, high-SR lower triangles indicate near-zero forgetting across all four suites, whereas for OOD the lower-triangular regions show a clear drop in performance: most cells are close to zero, while some isolated spikes of \emph{active cells} (under-diagonal cells with SR $> 0$) remain, revealing OOD forgetting.}
    \label{fig:classic_plus_heatmaps}
\end{figure}

We first reproduce the ID near-zero forgetting reported in \citet{liu2026pretrained}, as shown in Figure~\ref{fig:classic_plus_heatmaps}. The four evaluation matrices for the LIBERO suites show consistently high success rates (SR), indicating that the model retains strong performance while still learning new tasks. The NBT for the suites is slightly negative for Object ($-0.005$), Goal ($-0.017$), and LIBERO-10 ($-0.036$), and a small positive value for Spatial ($0.026$). These results show mild backward transfer for three of the suites and mild forgetting for Spatial, but driven only by one task, therefore confirming the high ID resistance to forgetting.

In contrast, the OOD evaluation on LIBERO-Plus shows predominantly low success rates, with most entries being exactly zero. Note that OOD performance is already low at task acquisition, with a mean SR fall from ID by about $0.27$, $0.27$, $0.25$ and $0.38$ on Spatial, Object, Goal and LIBERO-10 respectively, reflecting the initial domain shift (Appendix \ref{app:ood_acq}). However, we observe a substantial further degradation in performance after learning subsequent tasks. The mean OOD NBT is $+0.40\pm0.13$, $+0.48\pm0.17$, $+0.42\pm0.15$ and $+0.16\pm0.09$ on Spatial, Object, Goal and LIBERO-10 respectively (mean~$\pm$std.\ over the seven Plus categories), with the performance often collapsing to $\approx0$ after just one or two additional tasks.

Interestingly, the observed OOD catastrophic forgetting varies by suite: in Spatial and Object, the less complex suites, scattered cells retain non-trivial OOD performance, creating an irregular pattern of \emph{active cells} (under-diagonal cells with SR $> 0$). In the hardest suites, Goal and LIBERO-10, these patterns are mostly gone, leaving just a dead under-diagonal triangle.

\begin{figure}[t]
  \centering
  \includegraphics[width=\linewidth]{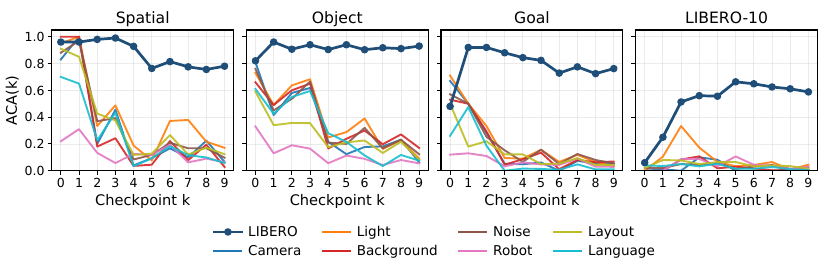}
  \caption{\textbf{Average Cumulative Accuracy over checkpoints for LIBERO
  and each LIBERO-Plus perturbation.} LIBERO ACA (blue) remains stable or
  increases slightly, while all LIBERO-Plus categories (colored lines) decline,
  showing a widening ID--OOD performance gap during training.}
  \label{fig:aca}
\end{figure}

Figure~\ref{fig:aca} plots the ACA over checkpoints. Despite the new irregular pattern of scattered active cells, OOD forgetting persists as new tasks are learned. The contrast with ID performance is clear: ID remains high and even slightly improves, indicating backward transfer in most suites, whereas OOD ACA declines continuously for most perturbations. This widening gap between LIBERO and LIBERO-Plus on the ACA shows how OOD robustness is rapidly eroded under continual learning, even while preserving in-domain performance.
\section{A Common Failure Pattern Across Perturbations}
\label{sec:architectural_bottleneck}

\subsection{The Behavioral Pattern}

We observed that OOD performance at acquisition drops for every perturbation, to a degree that depends on its type: lighting changes are the most benign, whereas changes to the robot's initial state are the most damaging (see Appendix~\ref{app:ood_acq}). Interestingly, however, we found that the perturbation type affects the magnitude of the acquisition drop but does not change the location of the \emph{active cell} spike pattern (see Section~\ref{sec:ood_forgetting})---these spikes occur at the same tasks and checkpoints.

To illustrate this active cell spike pattern, we construct, for each perturbation category, a vector containing $C(k,i,p)$ across all $i < k$ for a fixed $p$. We then compute the pairwise Pearson correlation $r$ between these vectors. For Spatial and Object, the correlations are strongly positive, with mean $\bar{r}=0.73$ and $0.82$, respectively, and all pairwise correlations are above $0.5$ (minimum $r=0.53$ for Spatial and $0.60$ for Object). Partial correlations given ID cell SR are nearly identical to the raw values (Spatial: $0.72$; Object: $0.81$), indicating that the cross-perturbation match is not an artifact of easy tasks staying easy. The full correlation matrices are shown in Appendix~\ref{app:extra_figs}. In contrast, for Goal and LIBERO-10, most under-diagonal cells are zero, and besides the complete forgetting, we observe no clear emerging behavioral pattern.

The behavioral pattern seen across perturbations for Spatial and Object suggests that OOD forgetting might stem from the model's internal representations rather than sensitivity to specific perturbations. If the OOD forgetting were driven, for example, by an inability to handle new camera angles, we would expect the camera perturbations to produce a distinct pattern from another category. Instead, the similar patterns across perturbations suggest a representation bottleneck in the ID distribution that limits adaptation to shifts across all domains in the same way.

\subsection{A Shared Representational Property}

To test whether the observed behavioral pattern reflects a property of the representation space rather than a coincidence, we ask if the structure of the representations is coherent across perturbations, as observed behaviorally with the active cells pattern. We use two metrics to check two complementary representational aspects. First, we look at the multimodal ID--OOD gap: we compute $\text{MMD}^2$ between the cells $C(k,i,\mathcal{D}_S)$ and $C(k,i,p)$ for a specific perturbation type $p \in \mathcal{P}_S$. Secondly, we measure multimodal drift with teacher-forced CKA between the checkpoint $\theta_i$ at which task $T_i$ is learned and every later checkpoint $\theta_k$ ($k>i$), on $T_i$ under domain $p$, i.e. between cells $C(i,i,p)$ and $C(k,i,p)$ (see Appendix~\ref{app:repr_settings}). For each suite, we flatten both metrics' per-category profiles into vectors---$\text{MMD}^2$ across all lower-triangular evaluation cells and CKA across all pairs $k>i$---and compute pairwise Pearson correlations across perturbation categories.

\begin{table}[h]
\centering
\caption{Cross-perturbation coherence: mean pairwise Pearson $\bar{r}$ between per-category flattened cells for multimodal ID--OOD $\text{MMD}^2$ and teacher-forced multimodal CKA. Positive values across all suites show that the shared structure extends to Goal and LIBERO-10, where behavioral coherence could not be estimated. Full correlation plots can be found in Appendix~\ref{app:extra_figs}.}
\label{tab:cell_coherence}
\renewcommand{\arraystretch}{1.2}
\begin{tabular}{lcccc}
\hline
\textbf{Suite} & Spatial & Object & Goal & LIBERO-10 \\
\hline
$\text{MMD}^2$ & 0.67 & 0.76 & 0.76 & 0.77 \\
\textbf{CKA}   & 0.64 & 0.75 & 0.72 & 0.84 \\
\hline
\end{tabular}
\end{table}

\begin{figure}[t]
  \centering
  \includegraphics[width=\linewidth]{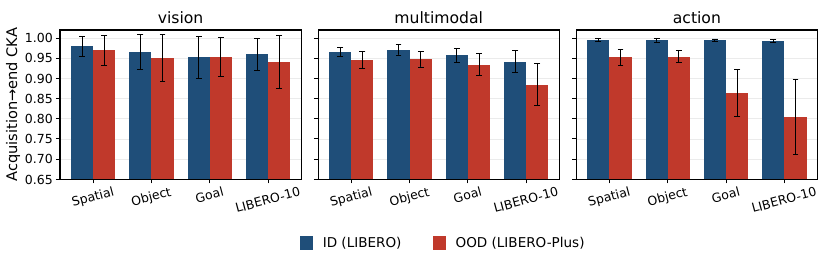}
  \caption{\textbf{Acquisition-to-end CKA drift by extraction module (vision, multimodal, action).} Each barplot shows mean $\text{CKA}(\theta_9,\theta_i)$ averaged over tasks, with the task std., for LIBERO (blue) and LIBERO-Plus (red). Vision is comparatively robust (small ID--OOD gap, within the task-to-task spread); multimodal shows a small but consistent gap; the action expert drifts the most, with higher variance on the harder suites.}
  \label{fig:cka_id_ood_drift}
\end{figure}

As illustrated in Table~\ref{tab:cell_coherence}, the representational metrics exhibit positive correlations across all suites, with no negative pairs in any perturbation combination. In other words, regardless of the perturbations being compared, the ID--OOD representational gaps and drift follow a similar structure. Notably, this pattern also holds for Goal and LIBERO-10, even where it is not apparent at the behavioral level because the performance collapsed to zero.

These results indicate that the pattern of cells showing larger representational gaps is shared across perturbation categories. While the perturbation type determines the overall magnitude and difficulty of overcoming the domain shift, it does not determine the relative profile of where the gap grows or shrinks across the evaluation matrix. This suggests that the observed structure stems from training dynamics and how the model represents the data, rather than from features specific to each perturbation.

\subsection{Module Localization of the Bottleneck}

In Figure~\ref{fig:cka_id_ood_drift}, we compare the mean acquisition $\text{CKA}(\theta_9,\theta_i)$ for tasks $T_i$ to measure how much the representation of a task drifts from acquisition to the end of continual fine-tuning. On the ID side, all modules exhibit low representation drift, consistent with the observed near-zero in-domain forgetting. In contrast, the OOD side shows different behaviors across extraction modules.

The vision encoder is comparatively robust: its ID--OOD gap is small (at most $0.02$ CKA) and within the task-to-task spread. The multimodal representation shows the first signs of consistent decay, with the OOD drift being higher than ID drift and, although the gap is small, it is systematic, low-variance, and consistent with the observed task difficulty. Finally, the action expert shows the largest gap between domains, but also higher variance, especially for Goal and LIBERO-10. Some of this noise can be explained by the fact that the action representation corresponds to the first of ten representations in the flow-matching process, and therefore can contain a significant portion of the noise.

\section{OOD Representations Misalign}
\label{sec:misalignment}

\begin{figure}[t]
    \centering
    \includegraphics[width=\linewidth]{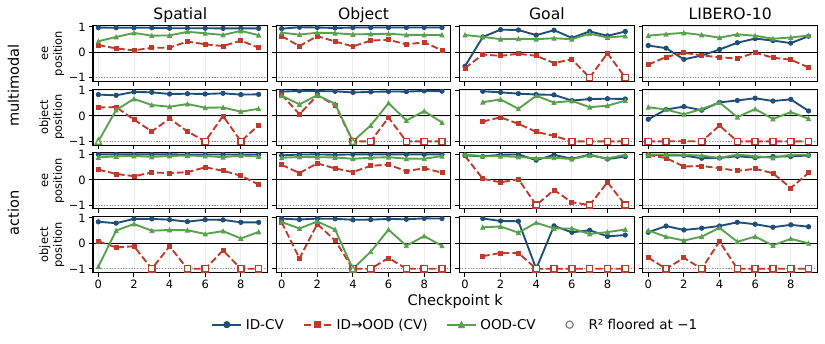}
    \caption{\textbf{Linear-probe $R^2$ for multimodal and action features
    predicting end-effector and manipulated object position.} Rows pair an extraction point with a regression target; columns are suites. Blue: ID-CV; red dashed:
    ID$\to$OOD transfer; green: OOD-CV. Open markers are floored at $-1$ because the $R^2$ regression metric can yield unbounded negative scores if it is worse than guessing the mean. Across all suites, ID$\to$OOD is consistently the worst score, while OOD-CV recovers, indicating that the target information remains linearly accessible out of distribution but the in-domain readout is misaligned.}
    \label{fig:probe_recovery}
\end{figure}

To obtain a first view of the information encoded in the model's hidden representations, we employed linear probes~\citep{alain2016understanding}. The probe's $R^2$ score quantifies how much of the target information can be linearly decoded from the representation. We fit linear probes on top of frozen features. For each checkpoint $\theta_k$ and task $T_i$, we extracted the pooled representations for multimodal and action extraction points, and set two regression targets: the 3D end-effector position, which serves as an easier baseline since it benefits from Pi0's proprioceptive state input, and the 3D position of the manipulated object, which cannot be directly inferred and provides the purer probe of visual-semantic readability.

To assess information loss, we ran three probe experiments: cross-validation trained and tested in ID (ID-CV), the same cross-validation-trained probes in ID but evaluated on LIBERO-Plus (ID$\to$OOD), and a cross-validated probe trained and tested on OOD (OOD-CV). This way we can respectively measure how much information is accessible in the original domain, the representation misalignment between ID and OOD, and the accessibility of the same information in OOD.

Figure~\ref{fig:probe_recovery} presents the regression linear probe evaluations. For the end-effector target, the pattern is clear for both multimodal and action features: the in-domain probe (ID-CV) achieves high $R^2$, while the transferred probes (ID$\to$OOD) show a sharp drop. The OOD-CV probes partially recover performance but do not regain the robustness of the original ID probe. Across all suites and checkpoints, the transferred probe is the worst of the three in $96.6\%$ of the cells for this target. The object-position target follows the same qualitative conclusion, holding for $92.9\%$ of the cells, though it is noisier. The median transfer $R^2$ for this target is around $-1.32$ (multimodal) and $-3.33$ (action), confirming the conclusion is not driven by a few easy cells, and both targets support the same pattern.

In general, the pattern is consistent: the transfer score is the worst, and although ID-CV dominates, OOD-CV recovers in most cases. This means that the information might still be accessible in the OOD data but misaligned, rather than a widespread loss of information.

\section{Discussion}
\label{sec:conclusion}

Near-zero forgetting is an important step toward practical continual learning in robotics, as it offers the possibility of avoiding the main cost of catastrophic forgetting and deploying policies that are robust under changing environments. In this work, we asked whether this regime still holds when already learned tasks are evaluated under domain shift. To do so, we reproduced the Pi0 experience-replay baseline of~\citet{liu2026pretrained} on the four LIBERO suites, reevaluated the same tasks under the seven LIBERO-Plus perturbations, and analyzed the resulting representations with CKA, MMD, and linear probes.

We found that despite near-zero forgetting in ID, the model incurs heavy OOD forgetting: as sequential fine-tuning progresses, the ID--OOD performance gap widens even while in-domain retention stays intact. We also found that the pattern of active cells---tasks that keep non-zero OOD success under a given checkpoint---is shared across the seven LIBERO-Plus categories for the suites that do not fully collapse, and that this shared structure shows up in the multimodal representations and potentially in the action representations, whereas the vision encoder remains comparatively robust. Finally, the linear probe experiments suggested that the information in the representations remains mostly accessible in the OOD regime but is misaligned, consistent with the model narrowing its representations to the specific ID data. These results not only reveal heavy OOD forgetting in this regime but also offer a potential explanation for the observed phenomena, such as the scattered active cell pattern: the structure of forgetting is shared across perturbations, which is more consistent with a common underlying cause than with perturbation-specific effects.

Overall, ID retention alone is not sufficient evidence of deployment readiness, since near-zero forgetting can coexist with heavy OOD forgetting. Technical challenges in how the model represents and structures the information still need to be addressed. Future studies in the same regimes should consider OOD-aware testing and mitigation strategies for the domain-shift challenges that appear in real-world scenarios.

\paragraph{Limitations.}

An important limitation is that we only evaluated one model, Pi0, in a single regime with a fixed replay strategy, and results come from a single seed. Our results are specific to this setting, and it is difficult to know how OOD robustness would look in other near-zero forgetting setups. Another important limitation is that the representational analysis is correlational, meaning the correlations found are consistent with causal relationships but not proven, since establishing causality would require controlled interventions to see how the model reacts. Finally, LIBERO-Plus has a non-uniform distribution of test scenes across tasks and perturbation categories, and although most cells have many samples, some contain very few scenes. This limitation is mitigated by focusing on aggregated metrics, but further discussion can be found in Appendix~\ref{app:small_n}.

\clearpage
\acknowledgments{
    This paper has received funding from the Flemish Government under the Methusalem Funding Scheme (grant agreement n° METH/24/009) and National Science Centre, Poland, grant no \mbox{2023/51/D/ST6/02846}. Bartłomiej Twardowski acknowledges the grant RYC2021-032765-I. 
}

%===============================================================================

\bibliography{IEEEabrv, references}  % .bib

\appendix
\clearpage

\section{LIBERO Suite Descriptions}
\label{app:libero_suites}

Table~\ref{tab:libero_suites} details the axes of variability for each of the four LIBERO suites referenced throughout the paper and sustains the claims of considering Spatial and Object easier suites with respect to Goal and LIBERO-10.

\begin{table}[h]
\centering
\caption{The four LIBERO benchmark suites and their axes of variability.}
\label{tab:libero_suites}
\footnotesize
\setlength{\tabcolsep}{3pt}
\renewcommand{\arraystretch}{1.15}
\begin{tabularx}{\columnwidth}{@{}l *{3}{>{\raggedright\arraybackslash}X}@{}}
\toprule
\textbf{Suite} & \textbf{Variability} & \textbf{Fixed Elements} & \textbf{Example} \\
\midrule
\textbf{Spatial} & Source & Environment, Action/Goal, Object, Destination &
  \textit{pick up the black bowl between the plate and the ramekin} \\
\addlinespace
\textbf{Object} & Object & Environment, Action/Goal, Source, Destination &
  \textit{pick up the alphabet soup and place it in the basket} \\
\addlinespace
\textbf{Goal} & Action/Goal, Object, Destination & Environment, Source &
  \textit{open the middle drawer of the cabinet} \\
\addlinespace
\textbf{LIBERO-10} & Environment, Action/Goal, Object, Source, Destination &
  --- & \textit{turn on the stove and put the moka pot on it} \\
\bottomrule
\end{tabularx}
\end{table}

\section{Training and Evaluation Settings}
\label{app:settings}

Table~\ref{tab:app_hyperparams} shows the hyperparameters used to fine-tune the model in the Continual Learning setup, which includes an Experience Replay buffer of $1000$ frames per task, $1$:$1$ current/replay mixing, LoRA on the VLM and action expert with full fine-tuning of the vision encoder and flow-matching projections. Training ran mainly on NVIDIA A100-SXM4-80GB GPUs; evaluation and representation rollouts mostly ran on A100-SXM4-80GB as well and, in some cases, on RTX 3090s.

\begin{table}[htb]
\centering
\caption{Continual fine-tuning hyperparameters (Pi0 / openpi low-memory LoRA recipe).}
\label{tab:app_hyperparams}
\small
\renewcommand{\arraystretch}{1.2}
\begin{tabular}{@{}ll@{}} 
\toprule
\textbf{Hyperparameter} & \textbf{Value} \\
\midrule
\multicolumn{2}{@{}l}{\textit{Architecture \& Compute}} \\
Base checkpoint & $\pi_0$ base pretrained \\
Precision & Mixed (bfloat16) \\
Hardware & NVIDIA A100-SXM4-80GB \\
\addlinespace

\multicolumn{2}{@{}l}{\textit{Optimization}} \\
Optimizer & AdamW ($\beta_1{=}0.9$, $\beta_2{=}0.95$, wd=$10^{-10}$) \\
LR schedule & Cosine (warmup $1000$, peak $2.5{\times}10^{-5}$, end $2.5{\times}10^{-6}$) \\
Schedule length & global $30{,}000$ decay steps \\
Steps per task & $10{,}000$ \\
Batch size & $8$ ($1$:$1$ ER mix within batch) \\
EMA & Disabled \\
\addlinespace

\multicolumn{2}{@{}l}{\textit{LoRA Configuration}} \\
VLM / Gemma & $r{=}16$, $\alpha{=}16$ \\
Action expert & $r{=}32$, $\alpha{=}32$ \\
\addlinespace

\multicolumn{2}{@{}l}{\textit{Environment \& Evaluation}} \\
ER buffer / mix & $1000$ frames/task; $\rho{=}0.5$ \\
Replan / settle & Every $5$ env steps / $10$ settle steps \\
ID eval episodes & $50$ fixed initial states / task \\
Eval horizons & Spatial: $220$, Object: $280$, Goal: $300$, LIBERO-10: $520$ \\
\bottomrule
\end{tabular}
\end{table}

Project page: \url{https://adrrp.github.io/vla-ood-robustness}.

\section{Task Sequences}
\label{app:task_order}

All suites use the task order of \citet{liu2026pretrained} (indices $0$--$9$). Table~\ref{tab:app_task_orders} lists the language instruction of each task; LIBERO-Plus evaluates the same underlying tasks under the seven perturbation categories.

\begin{table*}[!h]
\centering
\caption{LIBERO task order ($T_0$--$T_9$) per suite.}
\label{tab:app_task_orders}
\scriptsize
\setlength{\tabcolsep}{2pt}
\renewcommand{\arraystretch}{1.1}
\begin{subtable}[t]{0.48\textwidth}
\centering
\caption{Spatial}
\label{tab:app_tasks_spatial}
\begin{tabularx}{\linewidth}{@{}c >{\raggedright\arraybackslash}X@{}}
\toprule
\textbf{ID} & \textbf{Instruction} \\
\midrule
0 & pick up the black bowl between the plate and the ramekin and place it on the plate \\
1 & pick up the black bowl next to the ramekin and place it on the plate \\
2 & pick up the black bowl from table center and place it on the plate \\
3 & pick up the black bowl on the cookie box and place it on the plate \\
4 & pick up the black bowl in the top drawer of the wooden cabinet and place it on the plate \\
5 & pick up the black bowl on the ramekin and place it on the plate \\
6 & pick up the black bowl next to the cookie box and place it on the plate \\
7 & pick up the black bowl on the stove and place it on the plate \\
8 & pick up the black bowl next to the plate and place it on the plate \\
9 & pick up the black bowl on the wooden cabinet and place it on the plate \\
\bottomrule
\end{tabularx}
\end{subtable}
\hfill
\begin{subtable}[t]{0.48\textwidth}
\centering
\caption{Object}
\label{tab:app_tasks_object}
\begin{tabularx}{\linewidth}{@{}c >{\raggedright\arraybackslash}X@{}}
\toprule
\textbf{ID} & \textbf{Instruction} \\
\midrule
0 & pick up the alphabet soup and place it in the basket \\
1 & pick up the bbq sauce and place it in the basket \\
2 & pick up the butter and place it in the basket \\
3 & pick up the chocolate pudding and place it in the basket \\
4 & pick up the cream cheese and place it in the basket \\
5 & pick up the ketchup and place it in the basket \\
6 & pick up the milk and place it in the basket \\
7 & pick up the orange juice and place it in the basket \\
8 & pick up the salad dressing and place it in the basket \\
9 & pick up the tomato sauce and place it in the basket \\
\bottomrule
\end{tabularx}
\end{subtable}

\vspace{0.9em}

\begin{subtable}[t]{0.48\textwidth}
\centering
\caption{Goal}
\label{tab:app_tasks_goal}
\begin{tabularx}{\linewidth}{@{}c >{\raggedright\arraybackslash}X@{}}
\toprule
\textbf{ID} & \textbf{Instruction} \\
\midrule
0 & open the middle drawer of the cabinet \\
1 & put the bowl on the stove \\
2 & put the wine bottle on top of the cabinet \\
3 & open the top drawer and put the bowl inside \\
4 & put the bowl on top of the cabinet \\
5 & push the plate to the front of the stove \\
6 & put the cream cheese in the bowl \\
7 & turn on the stove \\
8 & put the bowl on the plate \\
9 & put the wine bottle on the rack \\
\bottomrule
\end{tabularx}
\end{subtable}
\hfill
\begin{subtable}[t]{0.48\textwidth}
\centering
\caption{LIBERO-10}
\label{tab:app_tasks_10}
\begin{tabularx}{\linewidth}{@{}c >{\raggedright\arraybackslash}X@{}}
\toprule
\textbf{ID} & \textbf{Instruction} \\
\midrule
0 & pick up the book and place it in the back compartment of the caddy \\
1 & put both moka pots on the stove \\
2 & put both the alphabet soup and the cream cheese box in the basket \\
3 & put both the alphabet soup and the tomato sauce in the basket \\
4 & put both the cream cheese box and the butter in the basket \\
5 & put the black bowl in the bottom drawer of the cabinet and close it \\
6 & put the white mug on the left plate and put the yellow and white mug on the right plate \\
7 & put the white mug on the plate and put the chocolate pudding to the right of the plate \\
8 & put the yellow and white mug in the microwave and close it \\
9 & turn on the stove and put the moka pot on it \\
\bottomrule
\end{tabularx}
\end{subtable}
\end{table*}

\section{Representation Metric Settings}
\label{app:repr_settings}

\subsection*{Feature Extraction}

To observe how the model's internal representations evolve, we compute different representation metrics at three feature-extraction points. The first is the vision representation, which is taken at the output of the vision encoder for each camera; the second and third are the multimodal and action representations. For these, all the observation data is fed with the initial action noise $A_t^0 \sim \mathcal{N}(0, I)$, and the final output is separated by prefix, corresponding to the multimodal representation (vision and language) and suffix, corresponding to the action representation taken just before it passes through the flow matching projection layers.

We lock the extraction at the flow-matching step $\tau = 0$ and run a single joint forward pass with all the data, as we do when computing the training loss. Because of the attention mask, the multimodal representation does not see the action tokens, and therefore this is identical to taking the final representation at the multimodal pass at inference.

The noise $A_t^0$ is the same for a given scene and frame across all checkpoints, and we use $\tau = 0$ because it is the only step where the action expert receives the same input in every checkpoint. At later steps the input depends on the velocities predicted by each checkpoint, so each model would be at a different part of the denoising process and the representations would not be comparable, even starting from the same noise. The downside is that the action representation at this step is closer to the noise, so we interpret it with caution. It is not just noise though, since the linear probes of Figure~\ref{fig:probe_recovery} are able to recover the object position from it.

To reduce to a single vector, we mean-pool all representations over the token axis. The unpooled sequence of tokens is a total of $N$ vectors, where each token $t_j \in \mathbb{R}^d$ has a feature dimension $d$. A binary mask $m_j \in \{0, 1\}$ is introduced to ignore the padding tokens to avoid merging with irrelevant information, and the tokens are averaged, obtaining a fixed-size vector $v \in \mathbb{R}^d$:

\begin{equation}
    v = \frac{1}{\sum_{j=1}^{N} m_j} \sum_{j=1}^{N} m_j t_j
\end{equation}

Specifically, for the vision representation, to distinguish the spatial identity of each camera view, since the number of cameras is constant, the tokens of each camera are pooled individually and then concatenated. The multimodal representation contains both the image and language tokens, which are pooled jointly into a single vector, and for the action representation the pooling is done across the $H$ action vectors without any masking.

\subsection*{Centered Kernel Alignment (CKA)}

With a sample of $n$ frames and a pooled feature dimension of $d$, the aligned feature matrices $X \in \mathbb{R}^{n \times d}$ and $Y \in \mathbb{R}^{n \times d}$ are used to compute Linear CKA~\citep{kornblith2019similarity}, which measures the structural similarity between two sets of representations. Given column-centered matrices $X_c$ and $Y_c$, CKA is defined as:

\begin{equation}
    \text{CKA}(X, Y) = \frac{\| Y_c^\top X_c \|_F^2}{\| X_c^\top X_c \|_F \cdot \| Y_c^\top Y_c \|_F}
\end{equation}

The CKA value lies between 0 and 1, with lower values showing completely different representations and higher values indicating an identical geometric structure.

We use CKA to study how a specific task $T_i$ evolves as the model learns new tasks. This procedure extracts and compares representations from two checkpoints, $\theta_k$ and $\theta_m$, evaluated on the same task data and domain $p$. In other words, we analyze how the representation varies in cells $C(\cdot, i, p)$.

Normally, evaluation is done in a closed-loop setup, where predicted actions define the robot's behavior and, therefore, the environment's next state. However, this raises a problem: checkpoints behave differently, so the observed environment will differ after just a few frames. To compare how the same data is treated across checkpoints, we need to use the same observations for both models.

Teacher Forcing does exactly that: instead of following the model's actions, a fixed set of expert trajectories is prerecorded from a reference checkpoint for all tasks $T_i$ and domains $p \in \{\mathcal{D}_S\} \cup \mathcal{P}_S$. These trajectories guide the robotic arm, ensuring identical environment perception for both models and guaranteeing frame-by-frame synchronization. For the expert trajectories of $\mathcal{D}_S$, we choose the final checkpoint after continual learning, $\theta_9$, for all tasks and suites. For the perturbed domains, we instead generate the trajectory for each task with the checkpoint trained on it, so for any task $T_i$ we record with $\theta_i$. This selection is done to ensure that both ID and OOD contain diverse trajectories, since there is no clear common set of checkpoints between ID and OOD that contain enough successful trajectories; the main diagonal is too low for some specific tasks of ID Spatial and LIBERO-10, and the last row fails catastrophically in OOD. We acknowledge that using a different reference checkpoint in each domain is a significant difference, even with the justification above. However, two observations suggest that our CKA conclusions are not driven only by this choice. First, the same frames are fed to the three extraction modules (vision, multimodal and action), and in Figure~\ref{fig:cka_id_ood_drift} the vision encoder shows a much smaller ID--OOD gap (at most $0.02$ CKA) than the multimodal and action modules, which would be unexpected if the trajectory source alone were driving it. Second, the cross-perturbation coherence of Table~\ref{tab:cell_coherence} also holds for $\text{MMD}^2$, which is computed in closed loop at a single checkpoint and does not use teacher forcing. Still, we interpret the magnitude of the CKA gap with caution.

We do not compute CKA between all frames; instead, whenever the model is triggered, which happens every 5 executed actions, and generates a new action chunk, we save the representations produced by that trigger. For a given scene rollout, we collect these periodically extracted representation vectors into a single feature matrix for each checkpoint. Then we compute CKA between the two scene-level matrices. When comparing two whole cells, we average the CKA produced by comparing each scene in the task that cell contains.

\subsection*{Maximum Mean Discrepancy (MMD)}

Maximum Mean Discrepancy~\citep{gretton2012kernel} is a kernel-based statistic that measures the distance between two distributions embedded in a reproducing kernel Hilbert space. Given samples $X = \{x_i\}_{i=1}^n \sim P$ and $Y = \{y_j\}_{j=1}^m \sim Q$, we estimate this distance with the unbiased $\text{MMD}^2$ estimator and an RBF kernel $k(x,y) = \exp\!\left(-\|x - y\|^2 / 2\sigma^2\right)$:

\begin{equation}
\begin{split}
    \text{MMD}^2(X, Y) &= \frac{1}{n(n-1)}\sum_{i \neq j} k(x_i, x_j) \\
    &\quad + \frac{1}{m(m-1)}\sum_{i \neq j} k(y_i, y_j) \\
    &\quad - \frac{2}{nm}\sum_{i,j} k(x_i, y_j)
\end{split}
\end{equation}

To set $\sigma$, we use the median of all pairwise Euclidean distances over the pooled set $X \cup Y$.

In population, $\text{MMD}(P,Q)$ is a metric between distributions in kernel space, so $\text{MMD}^2(P,Q)$ equals zero if and only if $P$ and $Q$ are indistinguishable under the RBF kernel. However, we report its unbiased finite-sample estimate from the formula above. Being unbiased, this estimate can have some noise when $P \approx Q$ (sampling noise), so we read values near zero as indistinguishable domains and larger positive values as a greater distributional gap.

For our specific case, we compute $\text{MMD}^2(k, i, p)$ to measure how far the representation of task $T_i$ departs from the ID manifold when it is evaluated under perturbation $p$ at checkpoint $\theta_k$. Because MMD compares sets of samples evaluated by the same policy, we do not need teacher forcing, so we measure MMD under normal closed-loop evaluation and directly compare the resulting pooled representation vectors of different scenes.

MMD assumes i.i.d. data, so comparing representations every time the model activates would make all frames strongly temporally correlated and violate the assumption. To mitigate the inherent temporal correlation of frames, each rollout only contributes with a fixed number of $K$ frames. Let $T$ be the number of total frames of a rollout, the number of sampled frames is defined by $K = \text{clip}(\lfloor T/20 \rfloor, 6, 12)$, which essentially ensures one frame per second of simulation. The 6 and 12 limits prevent longer episodes from disproportionately dominating the overall distribution.

\subsection*{Linear Probes}

Linear probes~\citep{alain2016understanding} are linear regressors trained on top of the frozen, extracted representations to predict a specific target variable. Because the probe is only a linear mapping, it serves as a diagnostic tool for identifying what information the representations linearly encode. Based on the probe's $R^2$ score, we obtain a quantitative measure of how much target information the representation contains.

In our case, the tested probe targets are the 3D positions of the manipulated object and the 3D end-effector (wrist). Since these are regression targets, $R^2$ is not bounded below because when the probe extrapolates, residual error can dominate total variance and drive the metric to large negatives. For practical reasons, when this happens, we floor and report the $R^2$ at $-1$, clearly indicating that the metric has been floored.

Linear probes use ordinary least squares with no $\ell_2$ penalty, evaluated with $5$-fold cross-validation grouped by episode/scene. The cross-validation partition is done by full episodes (scenes) rather than individual frames.

\section{OOD Acquisition per Perturbation Category}
\label{app:ood_acq}

Looking at OOD at acquisition (OOD-Acq) is the most reliable way to measure how the perturbations directly affect OOD performance because it is not influenced by the confounding effect of training on any posterior task. A drop from LIBERO to LIBERO-Plus then indicates a pure domain-shift penalty, disentangled from forgetting.

\begin{table}[ht]
\centering
\caption{OOD-Acq: mean diagonal success rate SR$(k,k)$ over the ten tasks, for LIBERO and each LIBERO-Plus category. Colored arrows show the change relative to LIBERO.}
\label{tab:performance_metrics}
\resizebox{\columnwidth}{!}{%
\begin{tabular}{l c c c c c c c c c}
\hline
Suite & LIBERO & \begin{tabular}{@{}c@{}}LIBERO-Plus \\ Mean\end{tabular} & Camera & Light & Bckg. & Noise & Robot & Layout & Lang. \\
\hline
Spatial & 0.79 & 0.52 & 0.46 & 0.86 & 0.51 & 0.61 & 0.26 & 0.54 & 0.41 \\
 & & \cellcolor{red!16}\small $\downarrow$0.27 & \cellcolor{red!20}\small $\downarrow$0.33 & \cellcolor{green!15}\small $\uparrow$0.07 & \cellcolor{red!17}\small $\downarrow$0.28 & \cellcolor{red!11}\small $\downarrow$0.18 & \cellcolor{red!32}\small $\downarrow$0.53 & \cellcolor{red!15}\small $\downarrow$0.25 & \cellcolor{red!23}\small $\downarrow$0.38 \\
\hline
Object & 0.91 & 0.64 & 0.68 & 0.87 & 0.74 & 0.76 & 0.21 & 0.61 & 0.57 \\
 & & \cellcolor{red!16}\small $\downarrow$0.27 & \cellcolor{red!14}\small $\downarrow$0.23 & \cellcolor{red!5}\small $\downarrow$0.04 & \cellcolor{red!10}\small $\downarrow$0.17 & \cellcolor{red!9}\small $\downarrow$0.15 & \cellcolor{red!42}\small $\downarrow$0.70 & \cellcolor{red!18}\small $\downarrow$0.30 & \cellcolor{red!20}\small $\downarrow$0.34 \\
\hline
Goal & 0.72 & 0.47 & 0.48 & 0.63 & 0.55 & 0.64 & 0.26 & 0.45 & 0.25 \\
 & & \cellcolor{red!15}\small $\downarrow$0.25 & \cellcolor{red!14}\small $\downarrow$0.24 & \cellcolor{red!5}\small $\downarrow$0.09 & \cellcolor{red!10}\small $\downarrow$0.17 & \cellcolor{red!5}\small $\downarrow$0.08 & \cellcolor{red!28}\small $\downarrow$0.46 & \cellcolor{red!16}\small $\downarrow$0.27 & \cellcolor{red!28}\small $\downarrow$0.47 \\
\hline
LIBERO-10 & 0.56 & 0.18 & 0.09 & 0.38 & 0.10 & 0.16 & 0.22 & 0.21 & 0.11 \\
 & & \cellcolor{red!23}\small $\downarrow$0.38 & \cellcolor{red!28}\small $\downarrow$0.47 & \cellcolor{red!11}\small $\downarrow$0.18 & \cellcolor{red!28}\small $\downarrow$0.46 & \cellcolor{red!24}\small $\downarrow$0.40 & \cellcolor{red!20}\small $\downarrow$0.34 & \cellcolor{red!21}\small $\downarrow$0.35 & \cellcolor{red!27}\small $\downarrow$0.45 \\
\hline
\end{tabular}%
}
\end{table}

Table~\ref{tab:performance_metrics} reports the average diagonal SR for each of the suites on ID evaluation, all the OOD perturbations, and the mean OOD-Acq. Domain shift generally imposes a consistent penalty across all suites, resulting in a significant drop in average performance. Notably, the relative degradation grows with baseline difficulty: the hardest in-domain suite, LIBERO-10, loses about $68\%$ of its ID success, versus $30$--$35\%$ in the other suites, consistent with poor model understanding of the tasks, which makes the harder OOD environment even more challenging.

\section{Low-Sample LIBERO-Plus Cells}
\label{app:small_n}

Within a suite, most evaluation tasks contain a healthy number of scenes, typically ranging from 30 to 50. However, the distribution of LIBERO-Plus test scenes is not uniform across tasks. A small subset of cells exhibits a lower sample count (10 or fewer scenes), occurring almost exclusively within the Background and Light perturbations. The minimum in the dataset is $N=4$ scenes per task, occurring only in four specific instances in the light perturbation (Spatial task 8, Object task 4, Goal task 1, LIBERO-10 task 2). Another few tasks in the background contain only 5 to 10 scenes, and Spatial task 0 for camera contains 6 scenes. Table~\ref{tab:app_small_n} enumerates them.

The variance these concrete cases can induce is largely mitigated for most metrics and results because we compare across tasks or perturbations and pool results. Where a numerical result isolates a single low-$N$ cell---for example, the Light ACA peak in LIBERO-10 around task 2, which has only 4 scenes---that spike should be read as a small-sample artifact rather than genuine recovery, and the localized uncertainty should be taken into account when interpreting it.

\begin{table}[h]
\centering
\caption{LIBERO-Plus evaluation cells with $N\le 10$ scenes.}
\label{tab:app_small_n}
\footnotesize
\setlength{\tabcolsep}{3pt}
\renewcommand{\arraystretch}{1.15}
\begin{tabular}{@{}llcc@{}}
\toprule
\textbf{Suite} & \textbf{Perturbation} & \textbf{Task} & \textbf{$N$} \\
\midrule
Spatial & Light & $8$ & $4$ \\
Object & Light & $4$ & $4$ \\
Goal & Light & $1$ & $4$ \\
LIBERO-10 & Light & $2$ & $4$ \\
Object & Background & $4$ & $5$ \\
Spatial & Camera & $0$ & $6$ \\
Spatial & Background & $8$ & $6$ \\
Goal & Background & $1$ & $6$ \\
LIBERO-10 & Background & $2$ & $8$ \\
Spatial & Background & $1$ & $10$ \\
Spatial & Light & $1$ & $10$ \\
\bottomrule
\end{tabular}
\end{table}

\section{Additional Behavioral, Coherence, and Probe Figures}
\label{app:extra_figs}

\begin{figure}
\centering
\includegraphics[width=\textwidth]{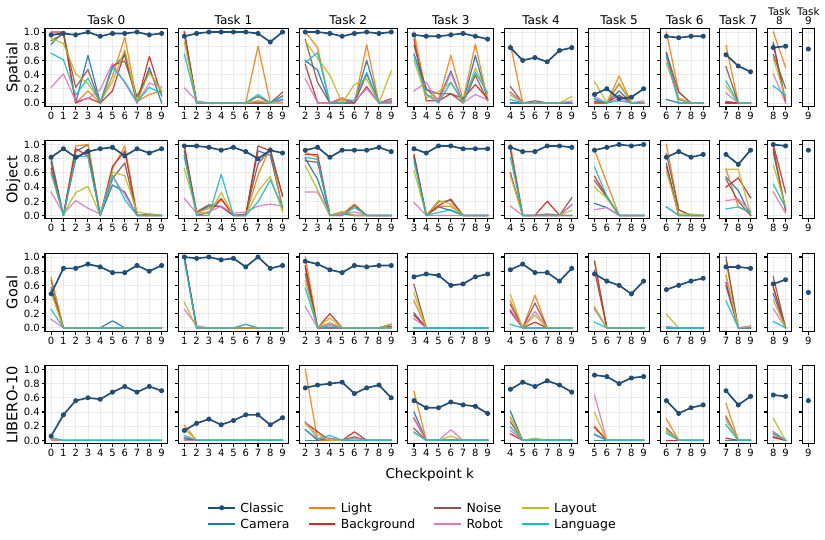}
\caption{\textbf{Per-task column trajectories of success rate under LIBERO
  and each LIBERO-Plus perturbation.} Each panel tracks task $T_i$ over checkpoints
  $k\ge i$ (panel width proportional to remaining checkpoints). Dark blue:
  in-domain LIBERO; other colors: the seven LIBERO-Plus categories. Complements the
  category-mean heatmaps in the main text by showing per-perturbation collapse
  and recovery along each task column.}
\label{fig:appendix_column_trajectories}
\end{figure}

In Section~\ref{sec:architectural_bottleneck}, we aggregate the LIBERO-Plus success rates over the perturbation categories. Figure~\ref{fig:appendix_column_trajectories} shows how each task column evolves under LIBERO and under every perturbation category in LIBERO-Plus individually.

\begin{figure}[h]
    \centering
    \includegraphics[width=\linewidth]{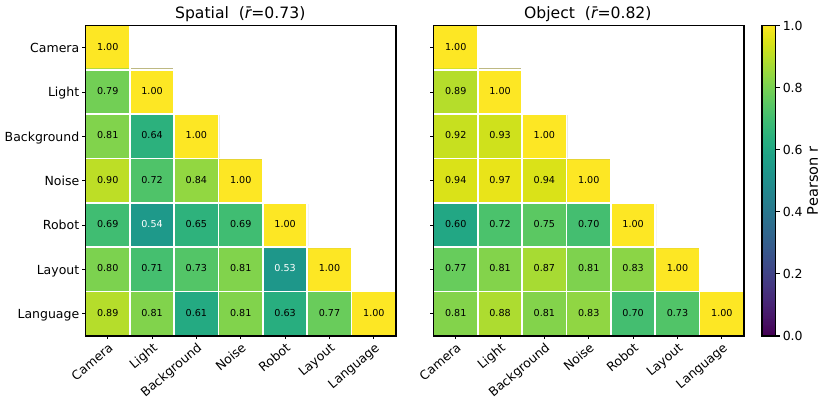}
    \caption{\textbf{Pairwise Pearson correlation of under-diagonal SR
    profiles across perturbation categories (Spatial and Object).} Each
    entry is the correlation between the flattened retention cells ($k > i$)
    of two LIBERO-Plus categories. The uniformly positive values mean that categories
    share the same pattern of active versus collapsed cells and differ mainly
    in scale.}
    \label{fig:perturbation_correlation}
\end{figure}

\begin{figure}
\centering
\includegraphics[width=\textwidth]{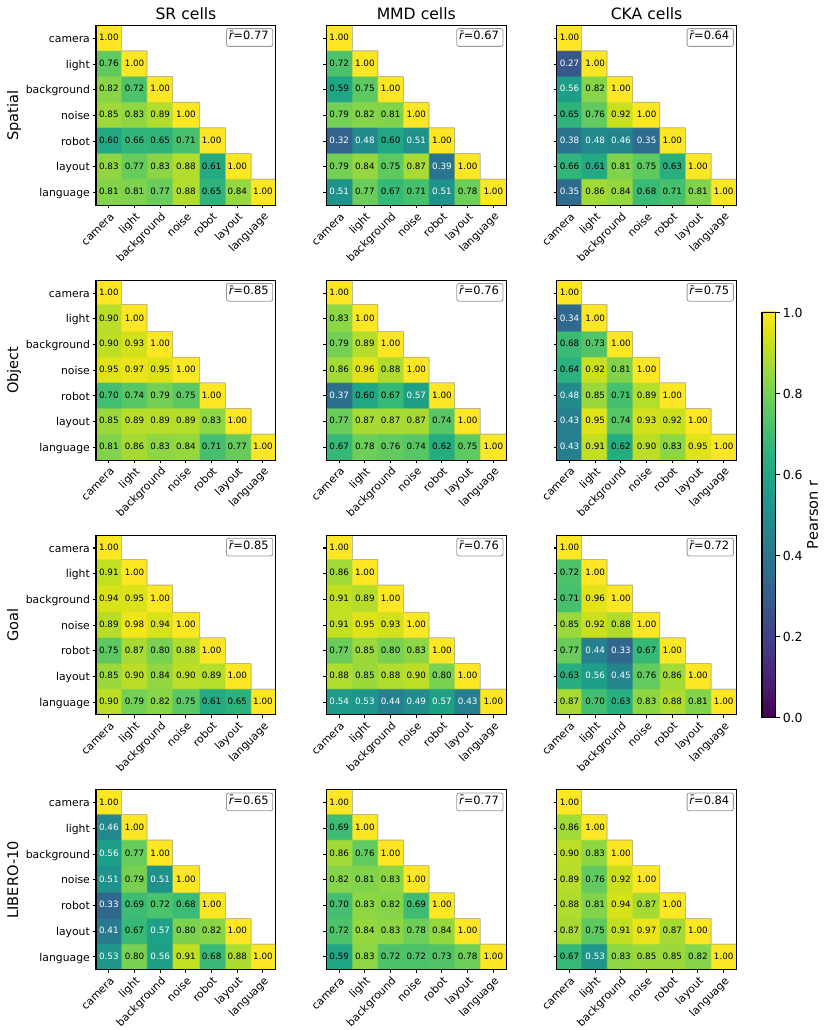}
\caption{\textbf{Full pairwise Pearson correlation of flattened lower-triangular cells across Plus categories.} Rows: suites; columns: success-rate (SR), multimodal ID--OOD $\mathrm{MMD}^2$, and multimodal CKA profiles. Each inset reports the mean pairwise $\bar{r}$. Unlike Figure~\ref{fig:perturbation_correlation}, which only uses the under-diagonal cells ($k > i$), the SR and $\mathrm{MMD}^2$ panels also include the diagonal ($k \ge i$), so the SR values differ slightly. The CKA panel uses the under-diagonal cells only ($k > i$), as it compares $\theta_k$ with $\theta_i$ on $T_i$, which is undefined on the diagonal. For Goal and LIBERO-10, most under-diagonal SR cells are zero, so their SR $\bar{r}$ is mainly driven by the diagonal and should be read with care.}
\label{fig:appendix_cell_coherence}
\end{figure}

We also reported the mean cross-perturbation coherence but did not show the full correlations; those can be found in Figure~\ref{fig:perturbation_correlation} (Spatial and Object SR profiles) and Figure~\ref{fig:appendix_cell_coherence} (full suite$\times$metric matrices).

\end{document}